%% file: root_new.tex
\documentclass[letterpaper, 10 pt, conference]{ieeeconf}  % Comment this line out if you need a4paper

\IEEEoverridecommandlockouts                              % This command is only needed if 
\usepackage{physics}
\usepackage[noadjust]{cite}
\usepackage{mathtools}
\usepackage{wrapfig}
\usepackage{amssymb}
\usepackage{flushend}

\usepackage{multirow}
\usepackage{makecell}
\usepackage{siunitx}

\usepackage{graphicx}
\usepackage{subcaption}
\let\labelindent\relax
\usepackage{enumitem}

\usepackage{newclude}
\usepackage{booktabs}
\usepackage{color}

\usepackage{soul}
\usepackage{algorithm}
\usepackage{algpseudocode}
\usepackage{url}

\newtheorem{conj}{Conjecture}

\title{\LARGE \bf
DEAR: Disentangled Environment and Agent Representations for Reinforcement Learning without Reconstruction}

\author{Ameya Pore$^{1}$, Riccardo Muradore$^{1}$ and Diego Dall'Alba$^{1}$% <-this % stops a space
\thanks{$^{1}$ Department of Engineering for Innovation Medicine, University of Verona, Italy}
\thanks{$\dagger$ Corresponding author: Ameya Pore (email: ameya.pore@univr.it)}}

\begin{document}

\maketitle
\thispagestyle{empty}
\pagestyle{empty}

%%%%%%%%%%%%%%%%%%%%%%%%%%%%%%%%%%%%%%%%%%%%%%%%%%%%%%%%%%%%%%%%%%%%%%%%%%%%%%%%
\begin{abstract}

Reinforcement Learning (RL) algorithms can learn robotic control tasks from visual observations, but they often require a large amount of data, especially when the visual scene is complex and unstructured. In this paper, we explore how the agent’s knowledge of its shape can improve the sample efficiency of visual RL methods. We propose a novel method, Disentangled Environment and Agent Representations (DEAR), that uses the segmentation mask of the agent as supervision to learn disentangled representations of the environment and the agent through feature separation constraints. 
Unlike previous approaches, DEAR does not require reconstruction of visual observations.
These representations are then used as an auxiliary loss to the RL objective, encouraging the agent to focus on the relevant features of the environment. We evaluate DEAR on two challenging benchmarks: Distracting DeepMind control suite and Franka Kitchen manipulation tasks. 
Our findings demonstrate that DEAR surpasses state-of-the-art methods in sample efficiency, achieving comparable or superior performance with reduced parameters. Our results indicate that integrating agent knowledge into visual RL methods has the potential to enhance their learning efficiency and robustness.
\end{abstract}

%%%%%%%%%%%%%%%%%%%%%%%%%%%%%%%%%%%%%%%%%%%%%%%%%%%%%%%%%%%%%%%%%%%%%%%%%%%%%%%%
\section{INTRODUCTION}

Visual perception is crucial for various robotics applications, such as autonomous driving, manipulation, and navigation.
While Reinforcement Learning (RL) algorithms have shown success in many of these tasks, learning control policies from high-dimensional and complex visual observations poses a significant challenge, which often requires a large amount of data and computational resources to achieve satisfactory performance \cite{levine2016end}. Moreover, real-world scenarios often involve dynamic and unstructured environments, where the visual scene can change significantly due to various factors, such as lighting, occlusion, or motion. These factors can introduce irrelevant or noisy information into the visual input, distracting or confusing the RL agent and hindering its learning process \cite{dunion2024conditional}.

To address this challenge, several encoding methods have been proposed to learn compact and informative image representations that can facilitate visual RL \cite{yarats2020image, yarats2021improving}. These low-dimensional feature vectors can then be used as the state input for various RL algorithms. However, most of these encoding methods are task agnostic, which can result in suboptimal or inefficient representations \cite{choi2023environment}. This can either miss or obscure the key features that are essential for the control task or include unnecessary features that are irrelevant or distracting for the RL agent.

Therefore, a key challenge for RL methods is to learn invariant representations that focus on the task-relevant aspects of the state and improve sample efficiency i.e. learning to remove distractors from an image can provide a good prior for vision-based RL \cite{zhang2021learning, mondal2022eqr, wang2021unsupervised}.
Previous methods proposed invariance based on reward through forward or inverse dynamics, i.e., two states are ``behaviorally equivalent" when they yield identical immediate rewards and similar subsequent state distributions \cite{zhang2021learning}; however, this approach does not perform well in challenging tasks \cite{dunion2023temporal}.
Another strategy is to learn disentangled representations of observations, which separate the informative factors of variation in an image from their underlying causes \cite{dunion2023temporal, dunion2024conditional}. However, these methods do not explain how the disentangled representations are interpretable.

\begin{figure}[t]
	\centering
	\includegraphics[width=0.9\linewidth]{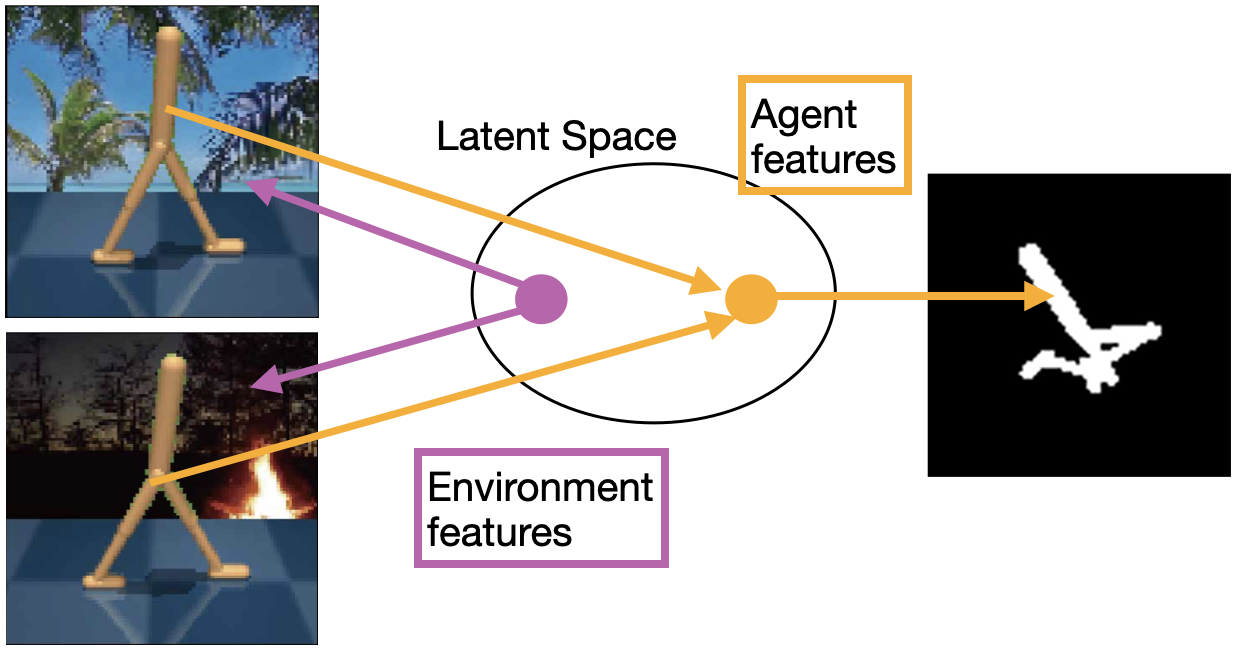}
	\caption{Robust representations of the visual scene should disentangle agent representations from the irrelevant environment information (e.g., changing background). Agent mask can extract agent representations from state observations, while representations not included in agent features can be pushed far apart.} \vspace{-4mm}
	\label{fig:top_image}
\end{figure}

One promising approach is to utilize the prior knowledge about an agent's physical characteristics, such as its morphology and joints, to learn invariant representations of the agent \cite{huang2020one}. This knowledge is generally practical, inexpensive, and known in advance. 
Hu et al. explored learning a factorized visual dynamics model that utilizes analytical forward kinematics of the robot and a learned environment transition model \cite{hu2022know}. 
While this approach can transfer the environment model to new robots with similar action spaces, 
it does not leverage the agent's knowledge during training. % \cite{hu2022know}.
Another direction is to implicitly train policies that consider an agent's morphology, using the policies' transferability as a signal \cite{dasari2021transformers}. Some methods directly train separate robot and task modules and attempt to transfer to new variations of the tasks \cite{pore2020simple}. All these studies learn self-perception in a self-supervised manner without extra agent information, which we assume to have in this study in the form of agent segmentation masks. We argue that explicitly learning agent representations could help in implicitly modeling environment representations, allowing visual RL approaches to learn faster and making the representations interpretable.

Recent studies have used agent masks as supervision to train self-perception modules to distinguish the agent and environment \cite{wang2021unsupervised, gmelin2023efficient}. 
Wang et al. proposed a method called VAI to learn visual agent masks from augmented input images, which are then used to feed only agent representations to a policy network \cite{wang2021unsupervised}. However, this method requires two pre-training steps for the auxiliary task, which can be computationally expensive. On the other hand, SEAR introduced a weakly supervised auxiliary loss, which is composed of the reconstruction loss of both the agent mask and the visual observation \cite{gmelin2023efficient}. 
This loss requires an additional decoder network to generate the original observation, and the encoded features overlap with the mask features, making it unclear how the environment features are disentangled. 

\begin{figure}[t]
	\centering
	\includegraphics[width=0.9\linewidth]{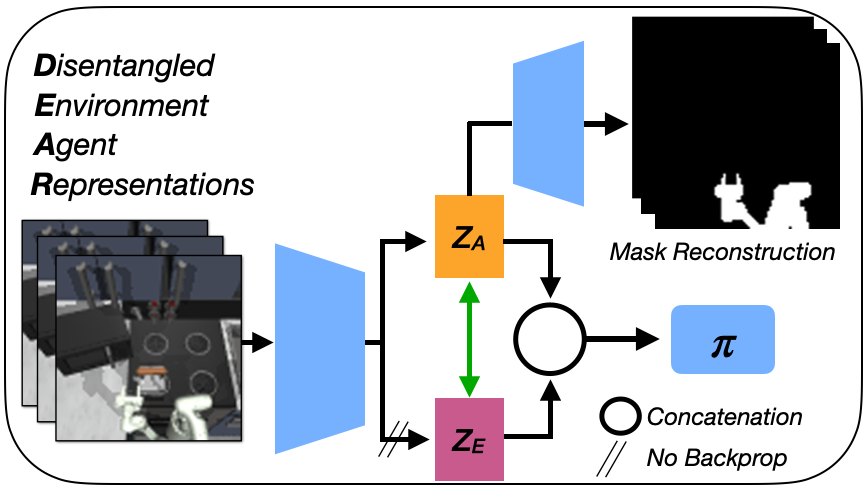}
	\caption{DEAR architecture: An agent mask is used as a supervision to encode agent features $z_A$. Additionally, the latent distance between $z_A$ and $z_E$ is maximized to encode environment features in $z_E$. A concatenation of $z_A$ and $z_E$ is used as the input to the RL policy $\pi$.
    } \vspace{-3mm}
	\label{fig:title_image}
\end{figure}

Instead of image reconstruction used in SEAR, we present Disentangled Environment and Agent Representations (DEAR), a method that reduces the overlap between the agent and environment representations (See Fig.~\ref{fig:top_image}). 
DEAR employs explicit objectives to effectively separate the agent and environment representations in the latent space, as depicted in Fig.~\ref{fig:title_image}. This is achieved through additive feature factorization, which enables the isolation of the agent features from the other features. 
We build on previous works that improve RL sample efficiency by incorporating additional losses into policy representations through self-supervised learning methods such as mask-based latent reconstruction \cite{yu2022mask}, contrastive learning \cite{laskin2020curl}, and effective image augmentation \cite{yarats2021improving}. DEAR introduces an auxiliary loss that maximizes the distance between the agent and environment representations in the latent space. 

We conduct experiments on two challenging benchmarks with diverse simulation environments: Distracting DeepMind (DM) control suite and Franka Kitchen manipulation tasks. DEAR exhibits improved sample efficiency compared to other baseline methods.
The analysis of the training parameters and times shows how DEAR obtains an optimal compromise between the complexity of the model and the training times while maximizing the latent distance between representations of the environment and the agent. To further evaluate the robustness of DEAR, we conducted ablation experiments with different auxiliary loss coefficients and noisy or approximate agent masks.

\section{Preliminaries}

We operate under the assumption that the environment is a fully observable Markov Decision Process (MDP). This is represented by a tuple $\mathcal{M} = (\mathcal{S}, \mathcal{A}, \mathcal{P}, \mathcal{R}, \gamma)$, where $\mathcal{S}$ is a set of states, $\mathcal{A}$ is a set of actions, $\mathcal{P}$: $\mathcal{S}$ × $\mathcal{S}$ × $\mathcal{A}$ $\rightarrow [0, 1]$ is the state-transition function, $\mathcal{R}$: $\mathcal{S}$ × $\mathcal{A}$ $\rightarrow \mathcal{R}$ is the reward function, and $\gamma \in [0, 1)$ is the discount factor. At any given time $t$, an RL agent selects an action $a_{t} \in \mathcal{A}$, based on its current state $s_{t} \in \mathcal{S}$ and its policy $a_{t} \sim \pi(s_{t})$. The agent then transitions to the next state according to the state-transition probability $P(s_{t+1}|s_{t}, a_{t})$ and receives a reward, $r_t = R(s_t, a_t)$. The goal of an RL agent is to learn a policy $\pi$ that maximizes the expected discounted cumulative rewards: $\mathrm{max}_{\pi} E_{P}[ \sum_{t=0}^{\infty} [\gamma^t r_{t}]]$. 
In image-based RL, the agent receives an observation of image pixels $X_{t} \in \mathcal{O} $ at time $t$, a high-dimensional representation of $s_t$. 

\begin{wrapfigure}{r}{0.20\textwidth}
  \begin{center}
    \includegraphics[width=0.19\textwidth]{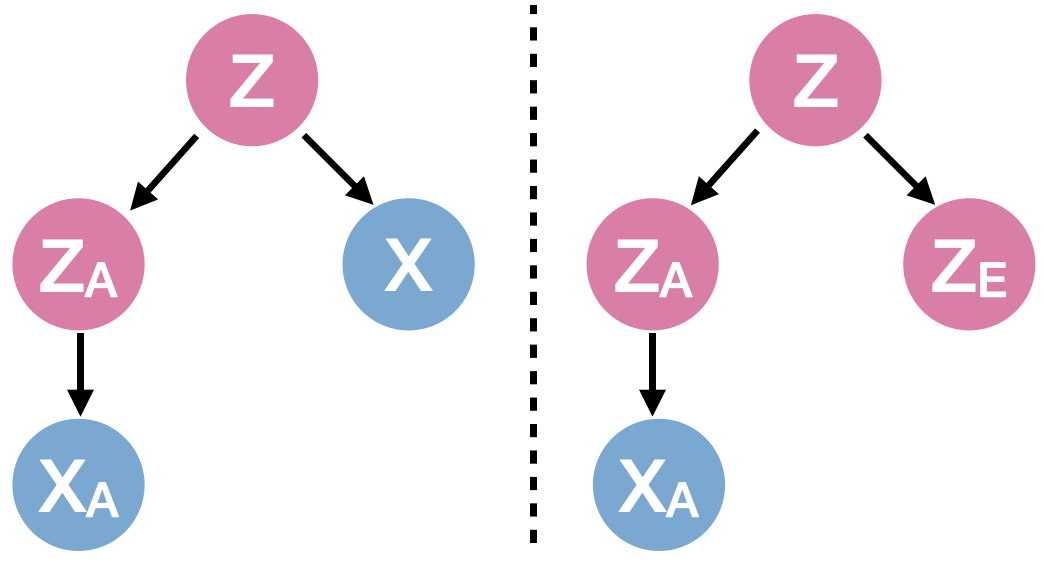}
  \end{center}
  \caption{Graphical model for (left) SEAR, and DEAR (right)}
  \label{fig:graphical}
\end{wrapfigure}

One of the approaches to learning effective control policies directly from raw image is to learn a low-level latent representation $z_{t} \in \mathcal{Z}$ of size $N << dim(\mathcal{O})$. As described in the previous section, learning such representations has been tackled in several ways: 1) injecting priors through data augmentation and 2) injecting auxiliary objectives through self-supervised learning. Data augmentation methods apply output-invariant perturbations to the labeled input example, such as random translate, crop, color jitter, etc. \cite{laskin2020reinforcement,yarats2020image}.  A common auxiliary loss is the reconstruction of the visual observation using autoencoders \cite{yarats2021improving}, which learn an encoder $q_\theta : \mathcal{O} \rightarrow \mathcal{Z}$ and a decoder $p_\phi : \mathcal{Z} \rightarrow \mathcal{O}$ by maximizing the log-likelihood of the image X.

\section{Learning disentangled representations for control}
We aim to investigate whether
1) agent mask information $X_A$ can be used to learn agent representations $z_A$ explicitly
2) agent representations $z_A$ can be used to model environment representations $z_E$ implicitly. 

\subsection{Disentangled Environment and Agent representations (DEAR)}

We begin by considering the graphical model from the previous approach \cite{gmelin2023efficient} (see Fig.~\ref{fig:graphical}), which aims to learn $z, z_A$ by maximizing $J = \mathrm{log}\ p(X, X_A)$ to learn a variational approximation $q_\theta(z,z_A| X)$.
We adapt the framework to suit our graphical model, where the goal is to learn $z_E$ and $z_A$. This is achieved by maximizing $J = \mathrm{log}\ p(X_E, X_A)$, where $X_E$ represents the environment image (excluding agent visuals). Assuming that $X_E$ and $X_A$ are conditionally independent at every timestep, we can rewrite $J$ as follows:
\begin{equation} \label{eqn:max}
J = \mathrm{log}\ p(X_E) + \mathrm{log}\ p(X_A)
\end{equation}
We state the following desired property
\begin{conj}
    \textit{If $z_A$ effectively encodes the agent information $X_A$ and $z_E$ effectively encodes environment information $X_E$, then the combination of $z_A$ and $z_E$ encodes the state representations $z$, i.e. $z = z_A + z_E$, and $z_A$ and $z_E$ are conditionally independent, i.e. $p(X_A | z_A) + p(X_E | z_E) = p(X | z) \leq 1$.}
\end{conj}

Using the conjecture in Equation.~\ref{eqn:max}, the evidence lower bound on J can be expressed as:
\begin{equation*} \label{eqn:derivation}
\begin{split}
    J & \geq E_{z_E \sim q_{\theta}} [\mathrm{log}\ p(X_E | z_E)] + E_{z_A \sim q_{\theta}} [\mathrm{log}\ p(X_A | z_A)] \\
     & = E_{z_A \sim q_{\theta}} [\mathrm{log}\ (1- p(X_A | z_A))] + E_{z_A \sim q_{\theta}} [\mathrm{log}\ p(X_A | z_A)] \\
     & \geq \mathrm{log}\ (1- E_{z_A \sim q_{\theta}}[p(X_A | z_A))] + \mathrm{log}\ (E_{z_A \sim q_{\theta}} [p(X_A | z_A)]) \\
     & \geq \mathrm{log}\ ( E_{z_A \sim q_{\theta}}[p(X_A | z_A)]) - \mathrm{log}\ (E_{z_A \sim q_{\theta}}[p(X_A | z_A)]^2) \\
     & = - \mathrm{log}\ ( E_{z_A \sim q_{\theta}}[p(X_A | z_A)]) \\
     & \geq - E_{z_A \sim q_{\theta}}[\mathrm{log}\ (p(X_A | z_A))] 
\end{split}
\end{equation*}
where $E_{z_A \sim q_{\theta}} [\mathrm{log}\ (p(X_A | z_A))]$ reconstructs the agent visual from $X_A$ and $z_A$. %Equation.~\ref{eqn:derivation} indicates that 
Under our assumptions, $p(X_A | z_A)$ is enough to model the environment features $z_E$ that maximize $J$. 

Previously, we explored how agent representations can be learned by having access to agent information. How to obtain agent-relevant information $X_A$?
We start by extracting the robot segmentation, denoted as $M$, from an image $X$ of the entire scene. 
The value of each pixel occupied by the robot is set to 1, while the rest is set to 0. A visual encoder, $q_{\theta}(.)$, is then used to generate a vector, $z$, from the image $X$. The vector $z$ is then split into two sub-vectors, $z_A$ and $z_E$. We feed the sub-vector $z_A$ into a decoder, $p_{\phi}(M|z_A)$, which predicts the robot mask. This decoder is trained using a Binary Cross-Entropy (BCE) loss:

\begin{equation} \label{eqn:mask_loss}
    \mathcal{L}_{mask} = M \mathrm{log}\ P_{\phi}(M|z_A) + (1-M)\ \mathrm{log}(1 - P_{\phi}(M|z_A))
\end{equation}

In order to learn the optimal values of $z_E$ and $z_A$, which are conditionally independent (according to Proposition 1), a feature separation constraint is enforced that minimizes their cosine similarity distance \cite{choi2023environment}: 

\begin{equation} \label{eqn:dis_loss}
    \mathcal{L}_{dis} = \abs{\norm{z_A}^{T}_2 \norm{z_E}_2}.
\end{equation}

This makes the representations orthogonal to each other. Note that our goal is not to reconstruct the observation state but rather to extract disentangled representations of $X$, which can be useful for the downstream RL control task, i.e., to capture features in $z_E$ that are independent of $z_A$.

\subsection{Visual RL using DEAR}
We use the above formulation to integrate DEAR in a visual RL algorithm.
The mask decoder and the disentanglement loss are added as auxiliary objectives, along with the RL algorithm loss:
\begin{equation} \label{eqn:loss}
    \mathcal{L} = \mathcal{L}_{RL} + \alpha \mathcal{L}_{mask} + \beta \mathcal{L}_{dis}
\end{equation}

The complete architecture diagram can be found in Fig.~\ref{fig:title_image}. 
The encoder and decoder structure for reconstructing the state observation proposed by Yarats et al. \cite{yarats2021improving} is used as an encoder and mask decoder, respectively, in our implementation.
The actor losses are not backpropagated into the encoder, and the data diversity is enhanced by applying random shifts to both robot masks and input image observations, following the example set by the DrQ-v2 algorithm \cite{yarats2021mastering}. Additionally, gradient flow is stopped at $z_E$ to prevent instability when learning $z_A$. 
Algorithm 1 provides a detailed explanation of the approach.

\input{algorithm}
\subsection{Value function Intuition}
In accordance with Proposition 1's graphical model, the value function can be expressed as $V(z) = V_A(z_A) + V_E(z_E)$. The value function dependent on the agent is represented by $V_A$, while $V_E$ represents the value function dependent on the environment. This formulation is capable of representing a broad range of value functions within the disentangled representation space.

Our main idea is that environmental characteristics can be captured by $V_E$, while the control of the robot to move the end effectors to a target location can be encoded in $V_A$. By explicitly modeling $z_A$, we believe that the agent can better understand both the environmental physics and its own control, which should result in faster learning.

Under our hypothesis, the agent should perform well in situations where the control of the agent is completely independent of the environmental changes, such as in distracting control environments \cite{stone2021distracting}. In these cases, the agent will only receive a reward for the robot's relevant motion, which is captured by $V_A$. However, in environments where the agent's actions significantly change the environment, the agent must first learn a good value function of the environment (i.e., $V_E$) before optimizing its behavior using $V_A$.
 
\section{Experiments} \label{sec:constrained_RL}

We evaluate DEAR in various challenging environments and tasks where the agent morphology ranges from a 2D ball and walker to a 7DoF robotic arm. All the environments used are based on the Mujoco engine. 
The experimental protocol is similar to the one followed in previous works \cite{laskin2020curl, laskin2020reinforcement, yarats2021mastering}
We first evaluate the agent on three different distracting DM control benchmarks \cite{stone2021distracting}, which is a variant of DM control with distractions added, and then two Franka kitchen environments for manipulating objects in a realistic kitchen with a Franka arm \cite{gupta2019relay}. Table.~\ref{Tab:env} provides a more detailed description of the considered environments. We report the average total reward for both benchmarks. 
\input{table}

\begin{figure}[t]
	\centering
	\includegraphics[width=\linewidth]{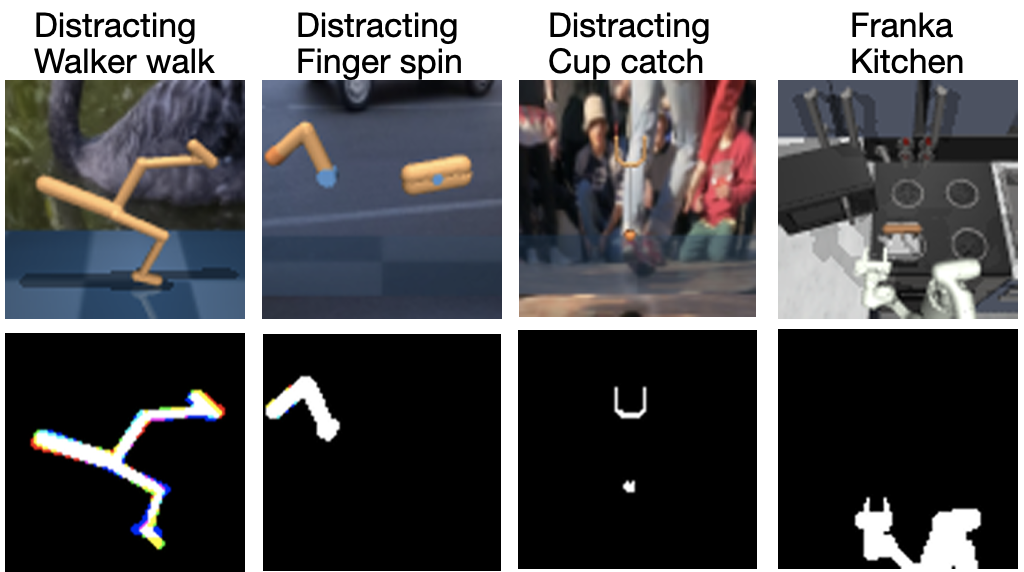}
	\caption{Different environments used for validation. (top row) RGB images that are input to the agent (bottom row) agent segmentation masks.
    }\label{fig:mask_image} \vspace{-4mm}
\end{figure}

The agent masks $M$ are obtained directly from the simulator, where ground-truth segmentation masks are generated using the mujoco rendering API. Examples of the environment frame with their masks are shown in Fig.~\ref{fig:mask_image}.
The agent receives an RGB image input $X$ and a mask $M$ to compute the supervised mask loss. Both the input and mask frames are 84x84 pixels. 

We use the same hyperparameters as optimized by the previous studies \cite{laskin2020reinforcement, yarats2021mastering}, with three new hyperparameters that DEAR introduces to the RL algorithm that includes a decoder learning rate and two coefficients for the mask and orthogonal loss ($\alpha$ and $\beta$ in Equation.~\ref{eqn:loss}). The value for the decoder learning rate is set the same as the critic learning rate to simplify hyperparameter tuning, while the values of $\alpha$ and $\beta$ are optimized using SEAR as a reference. More implementation details and hyper-parameters used can be found at the project website: \url{https://github.com/Ameyapores/DEAR}.

The performance of DEAR is evaluated against the following baselines:
\begin{itemize}[leftmargin=*]
    \item \textbf{DrQ-v2} \cite{yarats2021mastering}: Recognized as a standard baseline, DrQ-v2 employs DDPG with image augmentations for its simplicity and robust performance in visual RL tasks. It enhances data diversity by applying random shifts to input image observations, serving as an effective data augmentation technique.
    \item \textbf{TED} \cite{dunion2023temporal}: Adds a temporal classifier as an auxiliary loss to group frames closer in time and separate temporally distant frames. While we use this auxiliary loss with the DrQ-v2 RL loss, our method differs from TED as it does not consider the temporal aspect of RL observations. However, temporal information is also easy to access, providing an immediate self-supervision approach.
    \item \textbf{SEAR} \cite{gmelin2023efficient}: Utilizes agent masks for supervision via a mask loss and the original image for a reconstruction loss. These are added as auxiliary losses to the DrQ-v2 RL loss. Our method is closely associated with SEAR, especially in adopting a mask decoder. 
\end{itemize}

We carry out several analyses to evaluate the strengths of DEAR. 
\begin{enumerate}[leftmargin=*]
    \item \textbf{Disentanglement metrics}: The level of disentanglement is measured by comparing the cosine similarity distance between DEAR and the SEAR baseline.
    \item \textbf{Auxiliary objectives}: An ablation study was performed on the two auxiliary objectives within a distracting control task to determine the impact of these objectives on agent performance.
    \item \textbf{Optimal hyperparameters}: Different loss coefficients are tested to find optimal ones for our model.
    \item  \textbf{Robustness to Noise}: Generally, the mask is noisy and approximate in real-world conditions since it is extracted from real sensor data, subject to noise and artifacts. Thus, we artificially introduce noise in the mask by generating patches of pixels around the joints. A noise coefficient of 0.8 is used, meaning 80\% of noise is added. These masks were generated by randomly setting robot labels to non-robot labels with a user-specified probability. Additionally, an approximate mask with limited agent information is added (See Fig.~\ref{fig:noisy}). To generate an approximate mask, an original robot mask is downsampled and then upsampled back to its original size. Then, Gaussian blur and image thresholding are applied to get a new mask.
    
    \begin{figure}[t]
	\centering
	\includegraphics[width=1\linewidth]{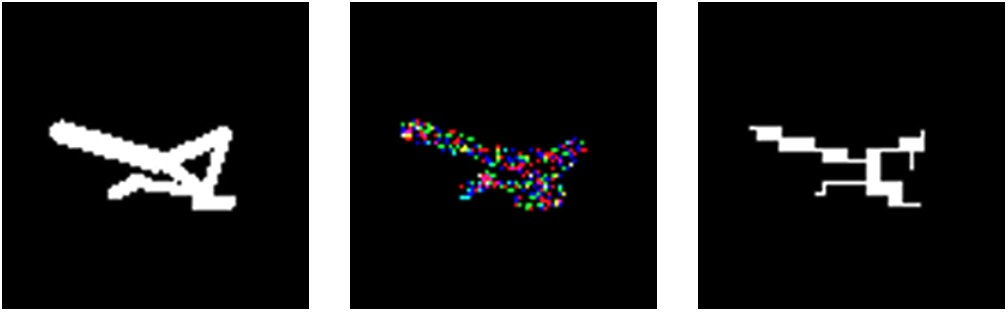}
	\caption{Different variants of agent segmentation masks used. From left to right: Normal segmentation mask, noisy mask, and approximate mask.
    }
	\label{fig:noisy}
\end{figure}
\end{enumerate}

\begin{figure*}[t]
	\centering
	\includegraphics[width=\linewidth]{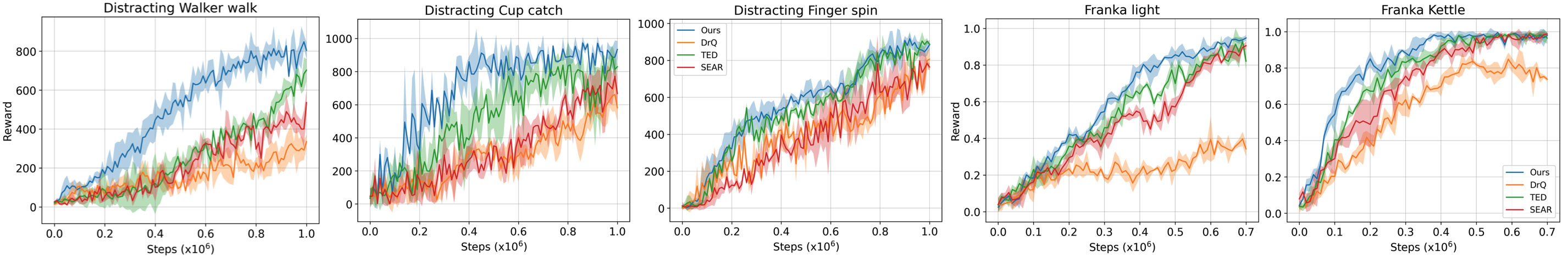}
	\caption{
    % Rewards for the three distracting DM control environments (left) Distracting Walker walk (middle) Distracting Cup catch (right) Distracting Finger spin. 
    Training curves of different methods on three distracting DM control environments and two Franka kitchen tasks. DEAR outperforms or matches other baselines on these tasks. Each curve is an average of 5 different seeds. \vspace{-3mm}
    }
	\label{fig:dmc_plot}
\end{figure*}

\section{Results}
\subsection{Benchmark control environment}
\textbf{Distracting Control Suite}:  
The results indicate that DEAR performs strongly compared to SEAR in all three environments and obtains a higher convergence for 1 million timesteps, increasing the sample efficiency by at least 25\%, shown in Fig.~\ref{fig:dmc_plot}. 
This suggests that DEAR can successfully disentangle representations of the environment that are not relevant to control. 

% \begin{figure}[t]
% 	\centering
% 	\includegraphics[width=\linewidth]{images/franka_plot.png}
% 	\caption{Training curves of different methods on two Franka kitchen tasks. 8 different seeds are used for each method.
%     }\label{fig:franka_plot}
% \end{figure}

\textbf{Franka Kitchen}: 
We see a superior performance of DEAR ($\sim 10\%$ improvement in sample efficiency) compared to other baselines on both tasks considered, shown in Fig.~\ref{fig:dmc_plot}, suggesting that even when the agent dynamically affects the environment, the agent can use the disentangled representations to improve the performance. We note that the margin of improvement shown by DEAR compared to DrQ-v2 is higher in the light switch task, which is a relatively more challenging task than the kettle.

In the distracting control suite, the agent explicitly follows the physical laws with high temporal correlations between visual observations. Therefore, TED can handle these scenarios optimally, performing better than SEAR. As for the Franka kitchen tasks, we believe that it is difficult to establish a strong temporal correlation due to the agent's interaction with the environment. As a result, TED and SEAR perform similarly on average in this case.
One advantage DEAR offers is feature interpretability, while for TED, it is hard to speculate which representations are considered important in improving performance. This opens up doors for future research to investigate the difference between the disentangled features of DEAR and TED.

\input{table2}

\textbf{Training parameters and time}: The training parameters and the wall clock training time per episode for each algorithm are shown in Table.~\ref{Tab:parameters}. DrQ-v2 uses a single visual encoder and, hence, consists of the least number of parameters. The time taken by DrQ-v2 is the minimum. On the contrary, SEAR uses two additional decoders, one for mask reconstruction and one for image reconstruction. Therefore, it has the maximum number of training parameters. SEAR, moreover, takes the maximum training time per episode. DEAR uses a single decoder for mask reconstruction, thereby requiring more parameters than DrQ-v2 but fewer than SEAR. Finally, TED requires an equivalent number of parameters to DrQ-v2 since the TED classifier consists of an additional 10k parameters. However, both DEAR and TED take an equivalent training time per episode. All the training experiments are carried out using a combination of RTX3080 or 1080Ti GPUs.

\textbf{Disentanglement metrics}: As observed in Table.~\ref{Tab:parameters}, 
we note that DEAR actively reduces the latent cosine similarity distance compared to SEAR. SEAR shows an overlap between the agent and environment vectors, likely contributing to its inferior performance relative to DEAR. Due to the different architecture of the other methods considered, calculating their latent cosine similarity distances is not feasible.

\subsection{Ablation study}
\begin{figure}[t]
	\centering
	\includegraphics[width=1\linewidth]{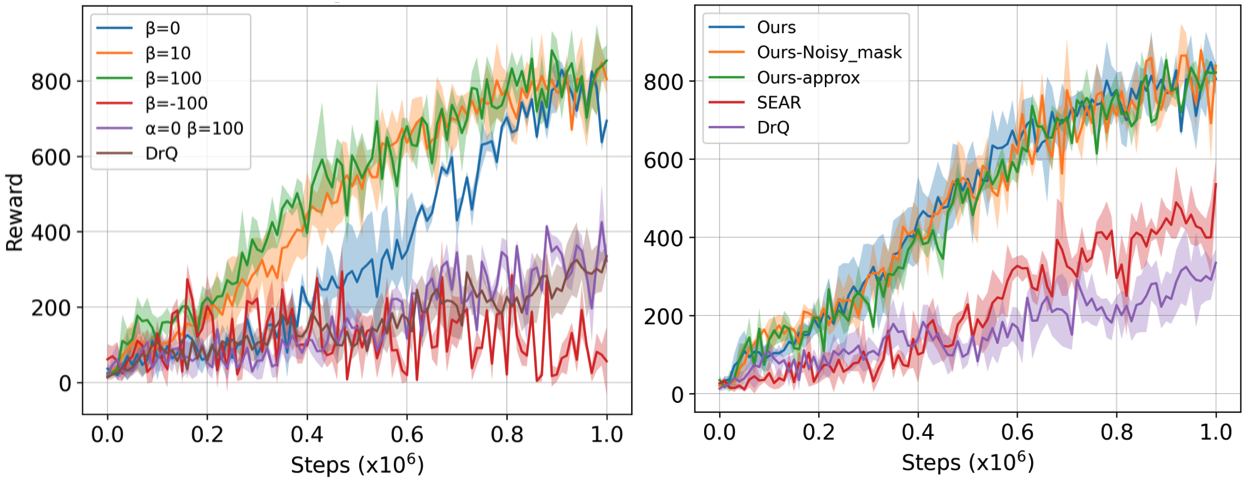}
	\caption{(Left) Experiments with different $\beta$ values provide insights into the contributions of the objectives in the overall loss function and also the changes with different hyperparameters. Where $\alpha$ values are not provided, $\alpha = 1$. Each experiment is an average of 5 random seeds. (Right) Training curves using noisy and approximate robot masks.
    } \vspace{-3mm}
	\label{fig:beta_plot}
\end{figure}

\textbf{Significance of objectives}: The results of the ablation study of the two auxiliary objectives are shown in Fig.~\ref{fig:beta_plot}.
The agent without any auxiliary objectives represents the DrQ-v2 algorithm. 
The agent trained with a single mask objective ($\beta=0$) produces superior performance to the one using a single disentanglement objective ($\alpha=0$), which produces an equivalent performance to the baseline DrQ-v2. This suggests that mask loss has a more substantial impact on the overall loss than the disentanglement loss. It is hard to interpret the generated representations when the mask objective is not considered. 
Finally, the best performance is obtained when the two objectives are used together.

\textbf{Robust to hyperparameters}: We provide experiments to identify the optimal coefficients that influence the performance in Fig.~\ref{fig:beta_plot} (left). We found $\beta = 100$ to be optimal for this task, and the results show some robustness for the values in the range $[1,100]$.  
Conversely, a negative coefficient leads to a reduced emphasis on disentanglement and the formation of overlapping vectors, which in turn diminishes performance. This indicates that weak disentanglement, characterized by highly overlapping vectors, hampers the agent's ability to differentiate between the environment and itself, adversely affecting performance.

\textbf{Robustness to noisy and approximate mask}: 

We see, from Fig.~\ref{fig:beta_plot}(Right), that the agent performance does not drop for both noisy and approximate masks, unlike SEAR, where a performance drop is reported \cite{gmelin2023efficient}. This robustness will be the subject for future research activities, 
extending the tests to experiments with real robotic setups. Recent advancements in segmentation models like SAM \cite{kirillov2023segment} could be leveraged as a readily applicable solution to produce the segmentation mask.

% \begin{figure}[t]
% 	\centering
% 	\includegraphics[width=0.9\linewidth]{images/noisy_mask.png}
% 	\caption{(Top row) different variants of agent segmentation masks used. From left to right: Normal segmentation mask, noisy mask, and approximate mask. (Bottom row) training curves using noisy and approximate robot masks.
%     }
% 	\label{fig:noisy_plot}
% \end{figure}

\section{CONCLUSIONS}  \label{sec:conclusions}

This study demonstrated the utility of disentangled representations in aiding RL agents with visual control tasks. By employing agent masks for supervision, we can encode distinct agent representations and enforce orthogonality with environmental representations, thereby minimizing their overlap. Our proposed technique seamlessly integrates with current RL frameworks, requiring only minor modifications. %to any encoder-based RL algorithms. 
Empirical evidence confirms that our approach surpasses conventional RL baselines in sample efficiency, validating the hypothesis that inductive bias through agent-specific masks can refine training.

One of the limitations of the current study is testing generalization to unseen environment parameters and adaptation to new tasks such as dynamic backgrounds, which will be extended in future works. Moreover, we will investigate if spatiotemporal disentanglement could further optimize sample efficiency by providing a stronger understanding of the general physics of the environment and the agent.

For real-world applicability, some new segmentation models such as SAM has shown promising performance on out-of-distribution robot/agent segmentation masks \cite{kirillov2023segment}. Using such models to generate the segmentation masks for supervision can show robust performance in real-world environments that are more noisier and dynamic than the controlled simulation settings. This translation from simulator to real would robotic agents will be part of upcoming works.



% \section*{ACKNOWLEDGMENT}

% Thank th support of xxx for this..

%%%%%%%%%%%%%%%%%%%%%%%%%%%%%%%%%%%%%%%%%%%%%%%%%%%%%%%%%%%%%%%%%%%%%%%%%%%%%%%%
% \clearpage
\bibliographystyle{IEEEtran}{

\bibliography{root}}

% \newpage
% Supplementary material

\end{document}

%% file: algorithm.tex
\begin{algorithm}
\caption{DEAR: Disentangled Environment and Agent Representations for control}\label{alg:dear}
\begin{algorithmic}[1]
% \function: UPDATECRITICANDDECODERS
% \mathbf{function} UPDATECRITICANDDECODERS(\mathcal{D})
% \:  ( x_t, m_t, a_t, R_(x_t, a_t), x_{t+n})
% \:  \mathcal{D} \leftarrow \mathcal{D} \cup ( x_t, m_t, a_t, R_(x_t, a_t), x_{t+n})
\item \textbf{for} $t = 1....T$ \textbf{do}
\item \ \: Collect transition $(x_t, M_t, a_t, R(x_t, a_t), x_{t+n})$
\item \ \: $\mathcal{D} \leftarrow \mathcal{D} \cup (x_t, M_t, a_t, R(x_t,a_t), x_{t+1})$ \Comment{Add the transition to Replay buffer $\mathcal{D}$ and update}
\item \ \: UpdateCriticandDecoder$(\mathcal{D})$
\item \ \: UpdateActor
\item \textbf{end for}
\item \textbf{function} UpdateCriticandDecoder
\item \ \:$(x_t, M_t, a_t, R(x_t, a_t), x_{t+n}) \sim \mathcal{D}$\Comment{Sample transition}
\item \ \: $z_t \leftarrow q_\theta(A_1(x_t))$ \Comment{Sample augmentation $A_1$}
\item \ \: $[z^{A}_t,z^{E}_t] \leftarrow z_t$
\item \ \: $\mathcal{L}_{mask} \leftarrow \mathcal{L}_{BCE}(P_{\phi}(z^{A}_t), A_1(M_t))$ \Comment{See Eqn.~\ref{eqn:mask_loss}}
\item \ \: Compute $\mathcal{L}_{dis} \leftarrow \mathcal{L}_(z^{A}_t,z^{E}_t)$ \Comment{See Eqn.~\ref{eqn:dis_loss}}
\item \ \: $\mathcal{L}_{total} \leftarrow \mathcal{L}_{critic} + \alpha\mathcal{L}_{mask} + \beta\mathcal{L}_{dis}$ \Comment{See Eqn.~\ref{eqn:loss}}
\item \ \: Update $\theta_{enc}, \theta_{critic}, \theta_{mask}$ using $\mathcal{L}_{total}$
\item \textbf{end function}
% \Ensure $y = x^n$
% \State $y \gets 1$
% \State $X \gets x$
% \State $N \gets n$
% \While{$N \neq 0$}
% \If{$N$ is even}
%     \State $X \gets X \times X$
%     \State $N \gets \frac{N}{2}$  \Comment{This is a comment}
% \ElsIf{$N$ is odd}
%     \State $y \gets y \times X$
%     \State $N \gets N - 1$
% \EndIf
% \EndWhile
\end{algorithmic}
\end{algorithm}

%% file: table.tex
\begin{table}
% \begin{normalsize}
\centering
\caption{Benchmark environments used}
\label{Tab:env}
\begin{tabular}{|p{1.2cm}|p{6.4cm}|} \hline
% \noalign{\smallskip }\hline\noalign{\smallskip}
% \cline{2-5} \cline{7-10} \noalign{\smallskip} 
Name & Description \\ \hline
% \noalign{\smallskip}\hline\noalign{\smallskip}
% Training parameters relative (M) & 0  & 4.24 & 0.1 & 2.12\\
Distracting Control Suite  & We choose three challenging environments from the distracting DM control suite: the Walker-walk, the Ball-in-the-cup, and the Finger spin, where a new background is randomly selected from the DAVIS dataset in each episode. \\ \hline
Franka Kitchen & Franka kitchen environments present tasks where the agent has to interact actively with the environment, unlike the DM control tasks. We choose the two most visually challenging single-goal tasks: turning the light switch and placing the kettle. These tasks are sparse reward settings where the agent receives a reward of 1.0 when it successfully turns the light switch to a pre-specified configuration or moves the kettle to a pre-specified goal position. \\ \hline
% \noalign{\smallskip}\hline
\end{tabular}
% \end{normalsize}
% \vspace{-15pt}
\end{table}

%% file: table2.tex
\begin{table}
% \begin{normalsize}
\centering
\caption{Number of training parameters, average wall clock training time per episode and latent distance between $z_A$ and $z_E$.}
% These experiments are carried out on an NVIDIA 1080Ti GPU.}
\label{Tab:parameters}
\begin{tabular}{ c c c c c }
\noalign{\smallskip }\hline\noalign{\smallskip}
% \cline{2-5} \cline{7-10} \noalign{\smallskip} 
Method & DrQ-v2 & SEAR & TED & DEAR\\
\noalign{\smallskip}\hline\noalign{\smallskip}
% Training parameters relative (M) & 0  & 4.24 & 0.1 & 2.12\\
Training parameters (M) & 12.86 & 16.21 & 12.87 & 14.09\\
Time per episode (s) & 12.55 & 18.02 & 16.19 & 16.11 \\
cos($z_A$, $z_E$) (1e-3) & - & 100 & - &1e-4 \\
\noalign{\smallskip}\hline
\end{tabular}
% \end{normalsize}
\vspace{-15pt}
\end{table}

%% file: root_new.bbl
\begin{thebibliography}{10}
\providecommand{\url}[1]{#1}
\csname url@samestyle\endcsname
\providecommand{\newblock}{\relax}
\providecommand{\bibinfo}[2]{#2}
\providecommand{\BIBentrySTDinterwordspacing}{\spaceskip=0pt\relax}
\providecommand{\BIBentryALTinterwordstretchfactor}{4}
\providecommand{\BIBentryALTinterwordspacing}{\spaceskip=\fontdimen2\font plus
\BIBentryALTinterwordstretchfactor\fontdimen3\font minus \fontdimen4\font\relax}
\providecommand{\BIBforeignlanguage}[2]{{%
\expandafter\ifx\csname l@#1\endcsname\relax
\typeout{** WARNING: IEEEtran.bst: No hyphenation pattern has been}%
\typeout{** loaded for the language `#1'. Using the pattern for}%
\typeout{** the default language instead.}%
\else
\language=\csname l@#1\endcsname
\fi
#2}}
\providecommand{\BIBdecl}{\relax}
\BIBdecl

\bibitem{levine2016end}
S.~Levine, C.~Finn, T.~Darrell, and P.~Abbeel, ``End-to-end training of deep visuomotor policies,'' \emph{The Journal of Machine Learning Research}, vol.~17, no.~1, pp. 1334--1373, 2016.

\bibitem{dunion2024conditional}
M.~Dunion, T.~McInroe, K.~S. Luck, J.~Hanna, and S.~Albrecht, ``Conditional mutual information for disentangled representations in reinforcement learning,'' \emph{Advances in Neural Information Processing Systems}, vol.~36, 2024.

\bibitem{yarats2020image}
D.~Yarats, I.~Kostrikov, and R.~Fergus, ``Image augmentation is all you need: Regularizing deep reinforcement learning from pixels,'' in \emph{International conference on learning representations}, 2020.

\bibitem{yarats2021improving}
D.~Yarats, A.~Zhang, I.~Kostrikov, B.~Amos, J.~Pineau, and R.~Fergus, ``Improving sample efficiency in model-free reinforcement learning from images,'' in \emph{Proceedings of the AAAI Conference on Artificial Intelligence}, vol.~35, no.~12, 2021, pp. 10\,674--10\,681.

\bibitem{choi2023environment}
H.~Choi, H.~Lee, S.~Jeong, and D.~Min, ``Environment agnostic representation for visual reinforcement learning,'' in \emph{Proceedings of the IEEE/CVF International Conference on Computer Vision}, 2023.

\bibitem{zhang2021learning}
A.~Zhang, R.~T. McAllister, R.~Calandra, Y.~Gal, and S.~Levine, ``Learning invariant representations for reinforcement learning without reconstruction,'' in \emph{Int. Conference on Learning Representations}, 2021.

\bibitem{mondal2022eqr}
A.~K. Mondal, V.~Jain, K.~Siddiqi, and S.~Ravanbakhsh, ``Eqr: Equivariant representations for data-efficient reinforcement learning,'' in \emph{International Conference on Machine Learning}.\hskip 1em plus 0.5em minus 0.4em\relax PMLR, 2022.

\bibitem{wang2021unsupervised}
X.~Wang, L.~Lian, and S.~X. Yu, ``Unsupervised visual attention and invariance for reinforcement learning,'' in \emph{IEEE/CVF Conference on Computer Vision and Pattern Recognition}, 2021, pp. 6677--6687.

\bibitem{dunion2023temporal}
M.~Dunion, T.~McInroe, K.~S. Luck, J.~P. Hanna, and S.~V. Albrecht, ``Temporal disentanglement of representations for improved generalisation in reinforcement learning,'' in \emph{The Eleventh International Conference on Learning Representations}, 2023.

\bibitem{huang2020one}
W.~Huang, I.~Mordatch, and D.~Pathak, ``One policy to control them all: Shared modular policies for agent-agnostic control,'' in \emph{International Conference on Machine Learning}.\hskip 1em plus 0.5em minus 0.4em\relax PMLR, 2020, pp. 4455--4464.

\bibitem{hu2022know}
E.~S. Hu, K.~Huang, O.~Rybkin, and D.~Jayaraman, ``Know thyself: Transferable visual control policies through robot-awareness,'' in \emph{International Conference on Learning Representations}, 2022.

\bibitem{dasari2021transformers}
S.~Dasari and A.~Gupta, ``Transformers for one-shot visual imitation,'' in \emph{Conference on Robot Learning}.\hskip 1em plus 0.5em minus 0.4em\relax PMLR, 2021, pp. 2071--2084.

\bibitem{pore2020simple}
A.~Pore and G.~Aragon-Camarasa, ``On simple reactive neural networks for behaviour-based reinforcement learning,'' in \emph{2020 IEEE International Conference on Robotics and Automation (ICRA)}.\hskip 1em plus 0.5em minus 0.4em\relax IEEE, 2020, pp. 7477--7483.

\bibitem{gmelin2023efficient}
K.~Gmelin, S.~Bahl, R.~Mendonca, and D.~Pathak, ``Efficient rl via disentangled environment and agent representations,'' 2023.

\bibitem{yu2022mask}
T.~Yu, Z.~Zhang, C.~Lan, Y.~Lu, and Z.~Chen, ``Mask-based latent reconstruction for reinforcement learning,'' \emph{Advances in Neural Information Processing Systems}, vol.~35, pp. 25\,117--25\,131, 2022.

\bibitem{laskin2020curl}
M.~Laskin, A.~Srinivas, and P.~Abbeel, ``Curl: Contrastive unsupervised representations for reinforcement learning,'' in \emph{International Conference on Machine Learning}.\hskip 1em plus 0.5em minus 0.4em\relax PMLR, 2020, pp. 5639--5650.

\bibitem{laskin2020reinforcement}
M.~Laskin, K.~Lee, A.~Stooke, L.~Pinto, P.~Abbeel, and A.~Srinivas, ``Reinforcement learning with augmented data,'' \emph{Advances in neural information processing systems}, vol.~33, pp. 19\,884--19\,895, 2020.

\bibitem{yarats2021mastering}
D.~Yarats, R.~Fergus, A.~Lazaric, and L.~Pinto, ``Mastering visual continuous control: Improved data-augmented reinforcement learning,'' \emph{arXiv preprint arXiv:2107.09645}, 2021.

\bibitem{stone2021distracting}
A.~Stone, O.~Ramirez, K.~Konolige, and R.~Jonschkowski, ``The distracting control suite--a challenging benchmark for reinforcement learning from pixels,'' \emph{arXiv preprint arXiv:2101.02722}, 2021.

\bibitem{gupta2019relay}
A.~Gupta, V.~Kumar, C.~Lynch, S.~Levine, and K.~Hausman, ``Relay policy learning: Solving long-horizon tasks via imitation and reinforcement learning,'' \emph{arXiv preprint arXiv:1910.11956}, 2019.

\bibitem{kirillov2023segment}
A.~Kirillov, E.~Mintun, N.~Ravi, H.~Mao, C.~Rolland, L.~Gustafson, T.~Xiao, S.~Whitehead, A.~C. Berg, W.-Y. Lo \emph{et~al.}, ``Segment anything,'' in \emph{Proceedings of the IEEE/CVF International Conference on Computer Vision}, 2023, pp. 4015--4026.

\end{thebibliography}
